\pdfoutput=1

\documentclass[11pt]{article}

\usepackage{ACL2023}

\usepackage{times}
\usepackage{latexsym}
\usepackage{subcaption}
\usepackage{graphics}
\usepackage{graphicx}
\usepackage{hyperref}
\usepackage{dsfont}
\usepackage{amsmath}
\usepackage{xspace}
\usepackage{verbatim}
\usepackage{xcolor}
\usepackage{soul}
\usepackage{graphicx}
\usepackage{tabularx}
\usepackage{booktabs}
\usepackage{caption}
\usepackage{geometry}
\usepackage{caption}

\usepackage[T1]{fontenc}

\usepackage[utf8]{inputenc}

\usepackage{microtype}

\usepackage{todonotes}

\usepackage{inconsolata}

\makeatletter
\def\blfootnote{\xdef\@thefnmark{}\@footnotetext}
\makeatother

\title{When Do Decompositions Help for Machine Reading?}

\author{Kangda Wei\textsuperscript{1}, Dawn Lawrie\textsuperscript{2}, Benjamin Van Durme\textsuperscript{2}, Yunmo Chen\textsuperscript{2*}, Orion Weller\textsuperscript{2*} \\ 
\textsuperscript{1}University of North Carolina Chapel Hill \\
\textsuperscript{2}Johns Hopkins University \\
\normalsize{\texttt{kangda@live.unc.edu,oweller@cs.jhu.edu}}
}

\begin{document}
\maketitle
\begin{abstract}
Answering complex questions often requires multi-step reasoning in order to obtain the final answer.
Most research into decompositions of complex questions involves open-domain systems, which have shown success in using these decompositions for improved retrieval. 
In the machine reading setting, however, work to understand when decompositions are helpful is understudied. 
We conduct experiments on decompositions in machine reading to unify recent work in this space, using a range of models and datasets.
We find that decompositions can be helpful in the few-shot case, giving several points of improvement in exact match.
However, we also show that when models are given access to around a few hundred or more examples, decompositions are not helpful (and can actually be detrimental).
Thus, our analysis implies that models can learn decompositions implicitly even with limited data.
\end{abstract}

\section{Introduction}\blfootnote{* Joint advising}
\label{sec:intro}
Much recent work has examined and improved models' ability to answer complex questions that require multiple steps of reasoning \cite{Wolfson2020Break,Dua2019DROPAR,yang2018hotpotqa,welbl-etal-2018-constructing,talmor-berant-2018-web}. A consistent theme in these works is to break the main complex question down into a series of sub-questions to be solved, which is referred to as question decomposition. These works generally represent decompositions as a human would, with explicit natural language sub-questions that build together to the final answer. Research has consistently shown that open-domain question answering (QA) models perform better on multi-step questions when they use decomposed sub-questions rather than answering with the main question alone \cite{Wolfson2020Break,perez2020unsupervised,geva2021did}.

\begin{figure}[t]
    \centering
    \includegraphics[width=\linewidth]{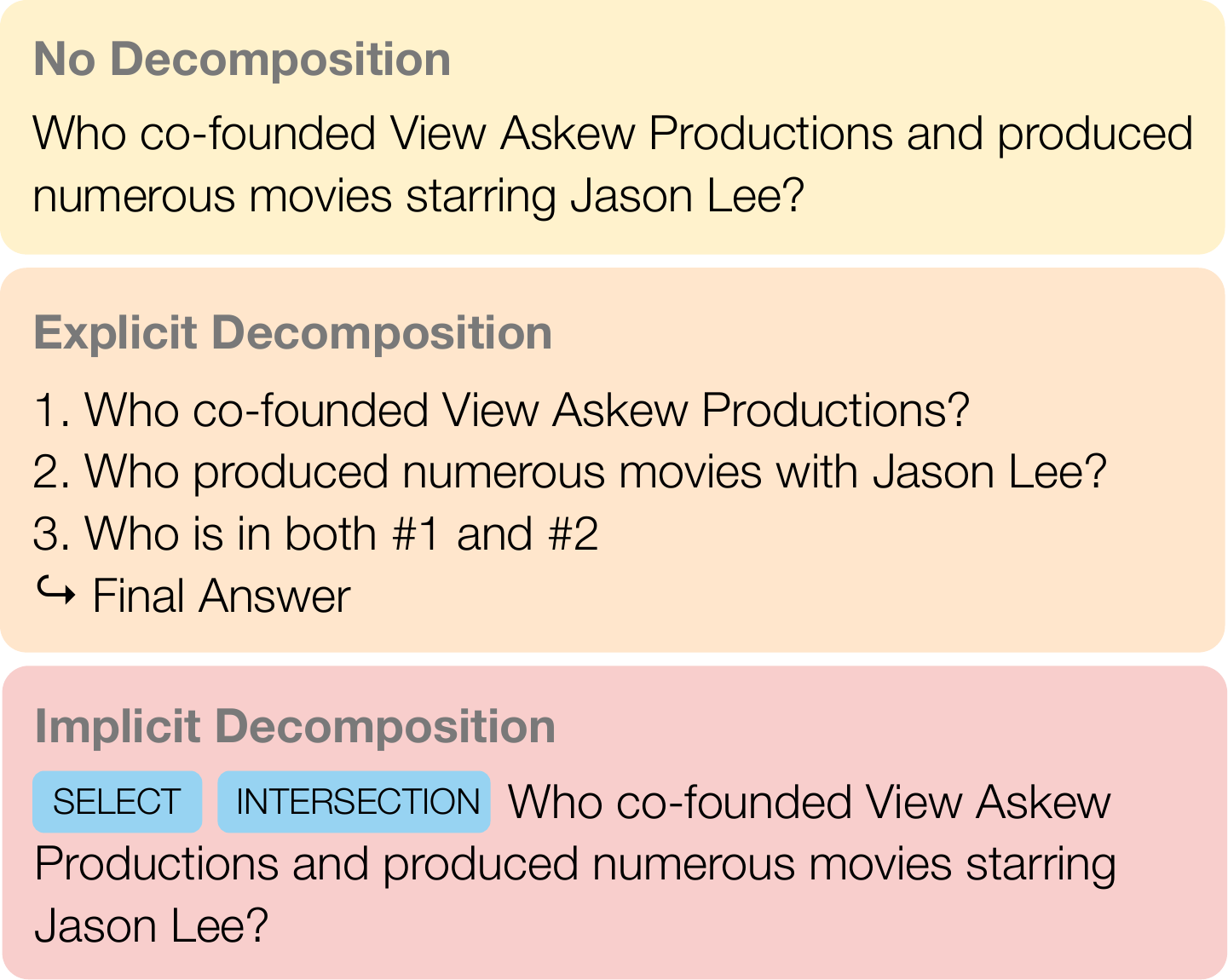}
    \caption{An instance of HotpotQA in BREAK \cite{Wolfson2020Break}, showing three different decomposition settings: (1) No Decomposition, i.e. regular question answering, (2) Explicit Decompositions that use iterative sub-questions, and (3) Implicit Decompositions that prepend the reasoning steps as special tokens.\vspace{-1.25em}}
    \label{fig:data}
\end{figure}

\begin{figure*}[t!]
\centering
\begin{subfigure}{.5\textwidth}
  \centering
  \includegraphics[width=.8\linewidth,trim={1.5cm 0.5cm 2cm 0cm}]{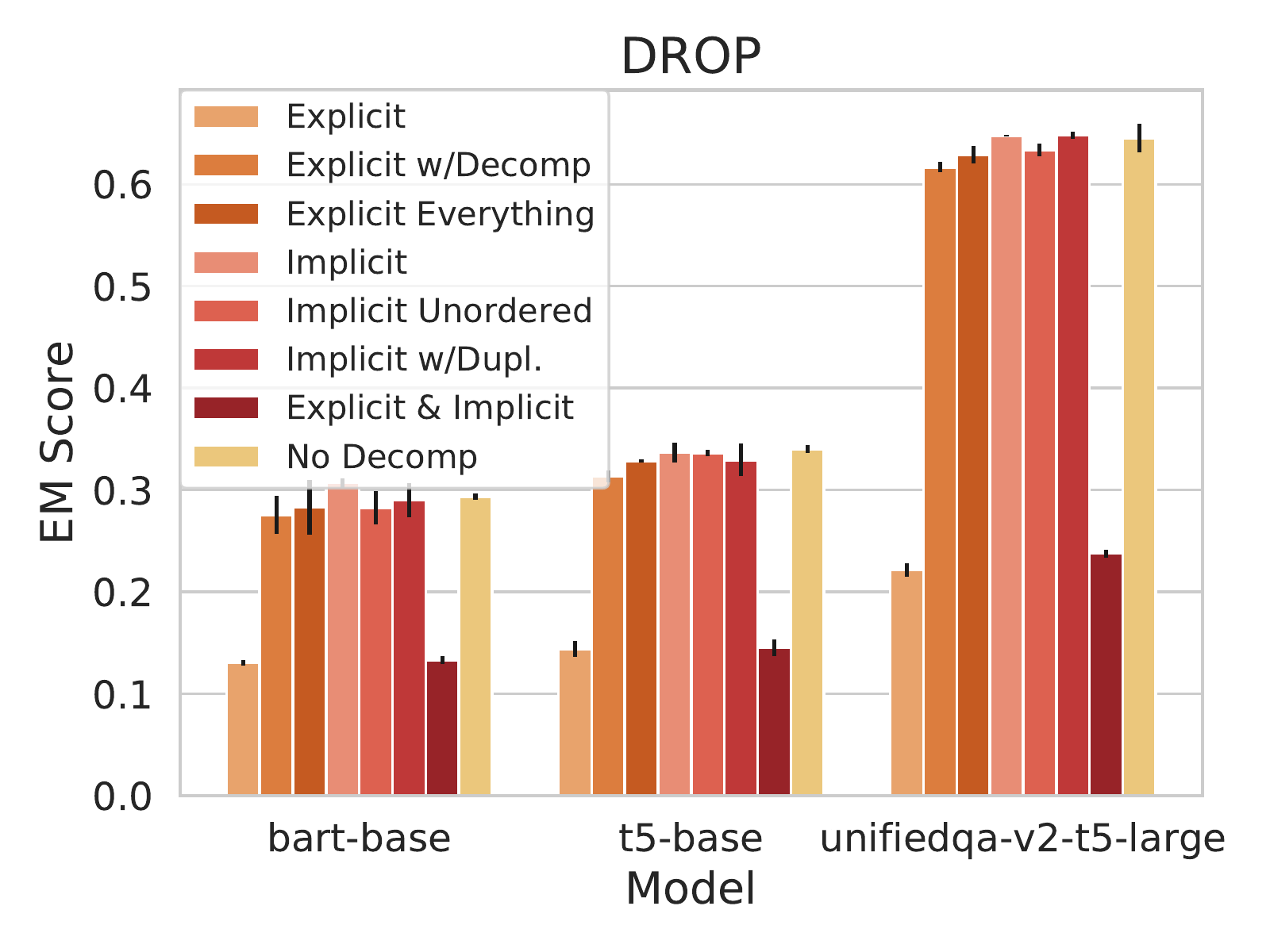}
  \label{fig:sub1}
\end{subfigure}%
\begin{subfigure}{.5\textwidth}
  \centering
  \includegraphics[width=.8\linewidth,trim={1.5cm 0.5cm 2cm 0cm}]{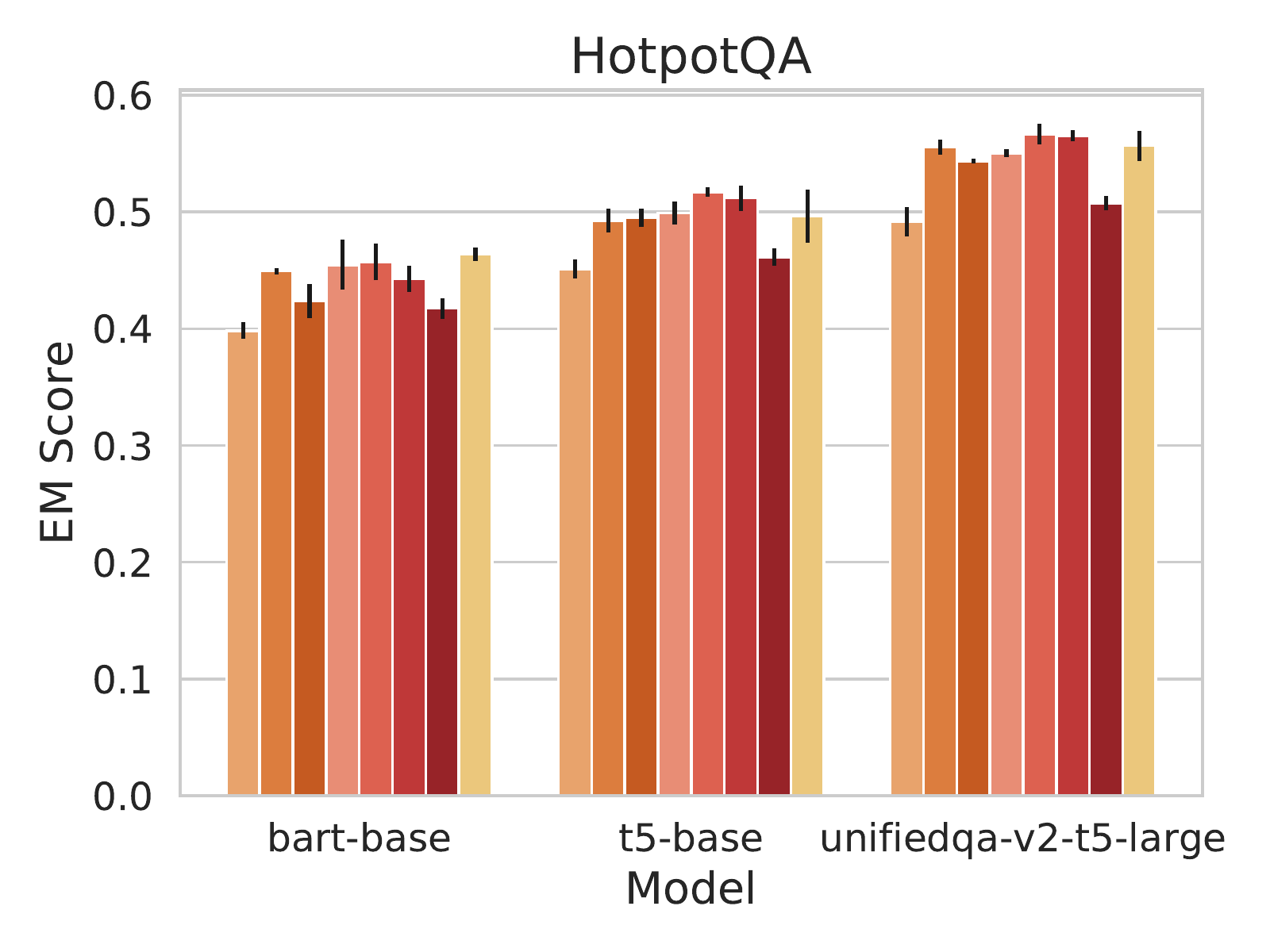}
  \label{fig:sub2}
\end{subfigure}
\caption{Main results showing the effect of various decomposition strategies. Left shows results on DROP, right shows HotpotQA. Runs were done over three random seeds and report the mean; error bars indicate 1 std. dev.\vspace{-0.75em}}
\label{fig:main}
\end{figure*}

Despite a large amount of research in decompositions for multi-step question answering, the majority of it has focused on using question decomposition for both information retrieval and question answering \cite{Wolfson2020Break,perez2020unsupervised}. The few works that have used decompositions for machine reading generally evaluate in limited settings \cite{guo2022complex,patel2022question}.

Therefore, we seek to shed light on \textbf{if and when} decompositions are helpful for machine reading. To do so, we analyze decomposition methods for seq2seq models across two multi-step QA datasets. Our results show that decompositions are only helpful in the low data setting, where there are less than a few hundred examples. Using decompositions in anything other than that setting performs the same (or much worse, depending on the strategy) than simply training the model end-to-end. 

Thus, overall, decompositions are helpful for question answering when they are used in two main settings: (1) information retrieval, where the decompositions help isolate distinct aspects of the question, or (2) in zero to low resource settings, where there isn't enough data to implicitly learn the multi-step process through end-to-end training.

\section{Experiment Setup}

\subsection{Data}
\label{sec:data}
We use the BREAK \cite{Wolfson2020Break} resource which contains annotated decompositions for 10 different benchmarks, including three reading comprehension benchmarks. We use two of these three datasets (HotpotQA and DROP) as the third, ComplexWebQuestions \cite{https://doi.org/10.48550/arxiv.1803.06643}, does not currently provide a training set.\footnote{Note that although the ComplexWebQuestions (CWQ) paper initially released the training set the authors have since removed it from the official Dropbox. Furthermore, even if the dataset was available CWQ does not verify that the questions are answered by the returned Google search.} Following BREAK and other follow-up work \cite{geva-etal-2021-aristotle}, we train and test our models using the high-level decompositions (also known as QDMRs).

Thus, we use HotpotQA~\cite{yang2018hotpotqa} and DROP \cite{https://doi.org/10.48550/arxiv.1903.00161} in our experiments, using the portions annotated from the train and validation sets. HotpotQA was created from workers on Mechanical Turk who annotated compositional questions from Wikipedia by using the page links (an example can be found in Figure~\ref{fig:data}). DROP was created to test discrete multi-hop reasoning and was also annotated by Mechanical Turk workers who used Wikipedia articles as the context. 

The BREAK annotations include the list of decomposed questions that eventually yield the same answer as the original question, along with an operator assigned to each decomposition that represents the type of reasoning for that step (e.g. Boolean, Comparison, etc.). Note that there are no gold labels for intermediate decomposition steps; the only ground truth label is for the main question.

\subsection{Models}
To explore the effect of decompositions on various types of common NLP models, we employ three different models: BART \cite{https://doi.org/10.48550/arxiv.1910.13461}, vanilla T5 \cite{https://doi.org/10.48550/arxiv.1910.10683}, and UnifiedQA-v2 \cite{khashabi2020unifiedqa,khashabi2022unifiedqa} that uses a t5 backbone and has been fine-tuned on other QA datasets but not on HotpotQA \cite{https://doi.org/10.48550/arxiv.1910.10683}. This allows us to demonstrate the effect of additional fine-tuning (UnifiedQA vs vanilla T5) as well as between different architectures (BART vs T5). Note that because UnifiedQA-v2 was multi-task trained on DROP, its scores are noticeably higher than the other models (Figure~\ref{fig:main}). However, our purpose is not to compare scores between models, but rather to compare scores \textit{between different decomposition strategies}. Thus, the inclusion of this model on DROP shows us that our results hold even if the model was pretrained on it. For more hyperparameter and compute details, see Appendix~\ref{app:details}.

\subsection{Decomposition Strategies}
There are many possible ways to combine decomposition with model fine-tuning. We try a wide variety of techniques (including novel ones) that we group into three categories: (1) no decomposition e.g. the baseline QA format, (2) explicit decomposition, and (3) implicit decomposition.

\paragraph{Explicit Decomposition} Explicit decompositions are the most common approach in the decomposition literature, generating the answer iteratively through sub-questions: the model answers the first decomposition step, then replaces placeholders in future decomposition steps with that predicted answer, then predicts the second decomposition step, and so forth. Note that using this method (\textit{Explicit}) naively presents issues with backpropagation, as the model can only backpropagate through the last decomposition step. Variations of strategies in this category include giving the model all previous decomposition steps as context (\textit{Explicit w/Decomp}) or including all decomposition steps and all predicted answers as context (\textit{Explicit Everything}).

\paragraph{Implicit Decomposition} Another way to do decompositions could be to add them implicitly. To do so, we utilize the \textit{operators} provided in the BREAK annotations which describe the type of reasoning needed, removing duplicate operators and keeping them in their given order. For example, in Figure~\ref{fig:data} the model uses \textit{select} twice and then \textit{intersection} in the explicit decomposition reasoning steps (but we remove the duplicate \textit{select}). We implement this in practice by adding a new special token for each operator and prepending them to the original question. Although this approach is novel in the context of decompositions, it bears similarity to work in the prompting literature, such as soft prompts \cite{qin-eisner-2021-learning,liu2022p}. Variations to this approach in the same category include randomizing the order of the special tokens (\textit{Implicit Unordered}),  leaving in duplicate special tokens (\textit{Implicit w/Dupl.}), or even prepending these operators to the explicit sub-questions in the Explicit Decomposition approach to combine the two strategies (\textit{Explicit + Implicit}).

\section{Results}
\paragraph{Full-Data Experiments} We see the main results in Figure~\ref{fig:main} with results for DROP on the left and HotpotQA on the right. Bars are colored-coded according to their method. All bars are the mean of three random seeds and error bars indicate the standard deviation. We see that most methods perform nearly the same, except for two that underperform: \textit{Explicit} and \textit{Explicit + Implicit}. Note that both of these have issues with training end-to-end, as an \textit{Explicit} decomposition is not differentiable through all decomposition steps. Thus, we only end up differentiating through the last step of the explicit decomposition steps, leaving the model unable to learn as effectively. All other approaches to decomposition perform comparably, given random seed variance (e.g. t5-base DROP \textit{Implicit Decomp.} is 33.7\% exact match \textpm\ 0.9\% vs \textit{No Decomp} 34.0\% \textpm\ 0.4\%). In fact, in this full data setting, the \textit{No Decomp} method performs better or statistically similar to every other method according to two-sample t-tests with the Bonferroni correction \cite{weisstein2004bonferroni}, across all datasets and models.

\begin{table*}[h]
\centering
\scalebox{0.8}{
\begin{tabular}{>{\raggedright}p{3cm}>{\raggedright}p{5cm}>{\raggedright}p{3.5cm}>{\raggedright}p{2.5cm}>{\raggedright}p{1.25cm}>{\raggedright\arraybackslash}p{2cm}}
 \toprule
 \textbf{Question} & \textbf{Decompositions} & \textbf{Content (shortened)} & \textbf{Intermediate Predictions} & \textbf{Answer} & \textbf{Error} 
 \\ [0.5ex] 
 \midrule
  Who was born first, Kwok Kin Pong or Edison Chen? & 
  \#1: when was Kwok Kin Pong born?\newline
  \#2: when was Edison Chen born?\newline
  \#3: which is the lowest of \#1,\#2? &
  Edison Koon-hei Chen (born 7 October 1980) .. Kwok Kin Pong (born 30 March 1987 in Hong Kong) .. & \#1. 7 October 1980  \newline \#2. 30 March 1987 \newline \#3: Kwok & Edison Chen & Wrong Prediction at Last Step \newline(18\%) \\ [0.5ex] 
 \midrule
 Are both Deerhunter and Nine Lashes American Christian rock bands? & \#1. is Deerhunter a American Christian rock band? \newline \#2. is Nine Lashes a American Christian rock band? \newline \#3: if both \#1 and \#2 are true & .. Nine lashes is an American Christian rock band .. Deerhunter is an American rock band from Atlanta .. & \#1. Yes \newline \#2. Yes \newline \#3. Yes & No & Error propagation \newline(40\%) \\ [0.5ex] 
 \midrule
  What actor from ``Willow" also starred in ``The Usual Suspects"?& 
  \#1: who is the actor that starred in The Usual Suspects? \newline\#2: \#1 that was a actor from Willow? &
  .. Kevin Elliot Pollak .. a role in "Willow" .. the Usual Suspects stars Kevin Pollak .. & \#1. Kevin Pollak\newline \#2. Kevin Pollak & Kevin Elliot Pollak & Invalid or Missing Annotation \newline(42\%) \\[1ex] 
 \bottomrule
\end{tabular}}
\caption{Error analysis for standard decomposition (i.e. \textit{Explicit Decomp} with no fine-tuning, a la \citet{patel2022question}) on HotpotQA. Percentages calculated from annotation of 50 instances with representative examples shown.\vspace{-0.5em}}
\label{table:error analysis}
\end{table*}

\begin{figure}[!t]
    \centering

\begin{subfigure}{0.5\textwidth}
 \includegraphics[scale=0.5, trim={0.7cm 0.25cm 0cm 0cm}]{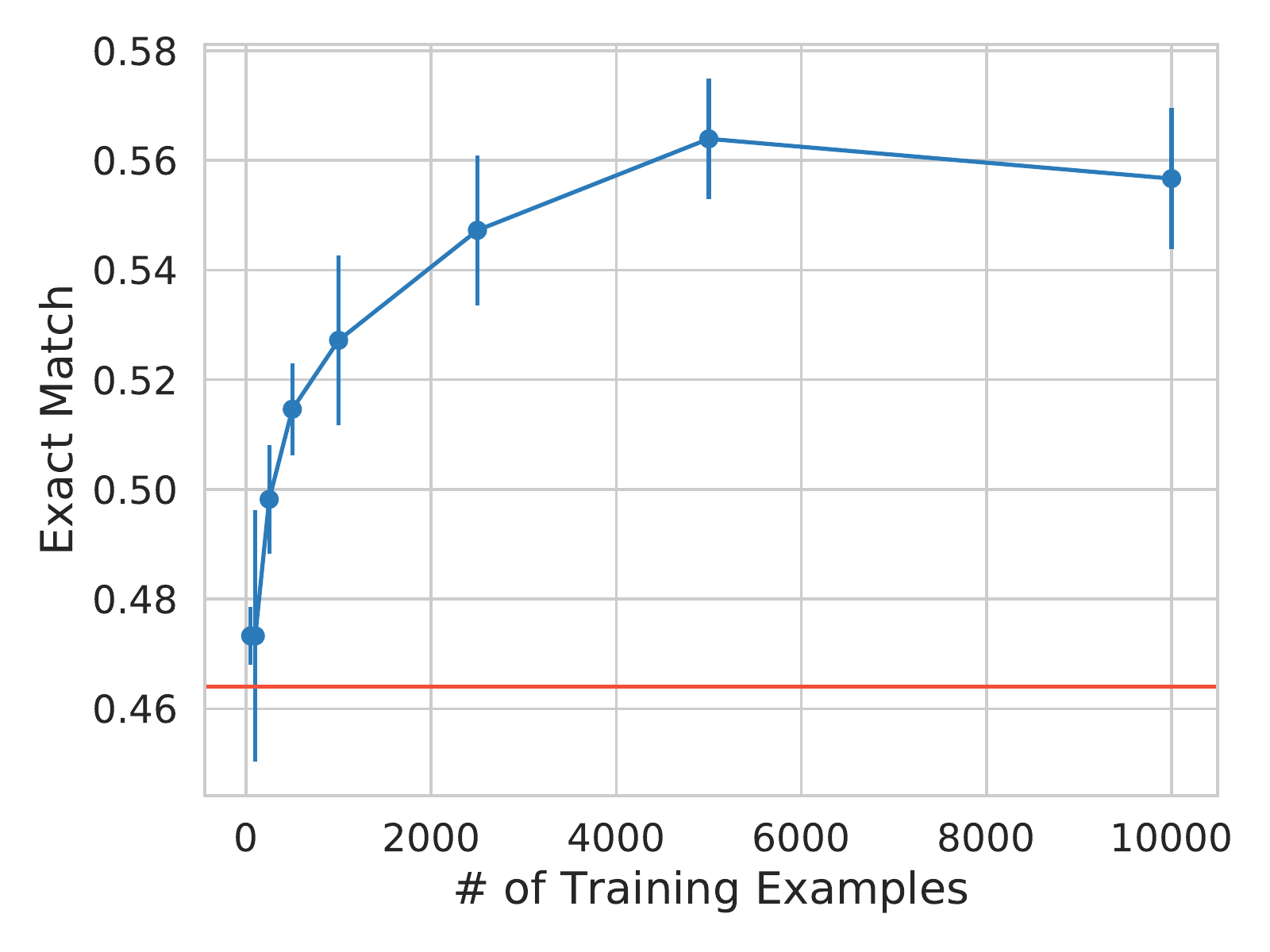}
    \caption[]{Results from UnifiedQA-v2\label{fig:size_unified}}
\end{subfigure}

\begin{subfigure}{0.5\textwidth}
 \includegraphics[scale=0.5, trim={0.7cm 0.25cm 0cm 0cm}]{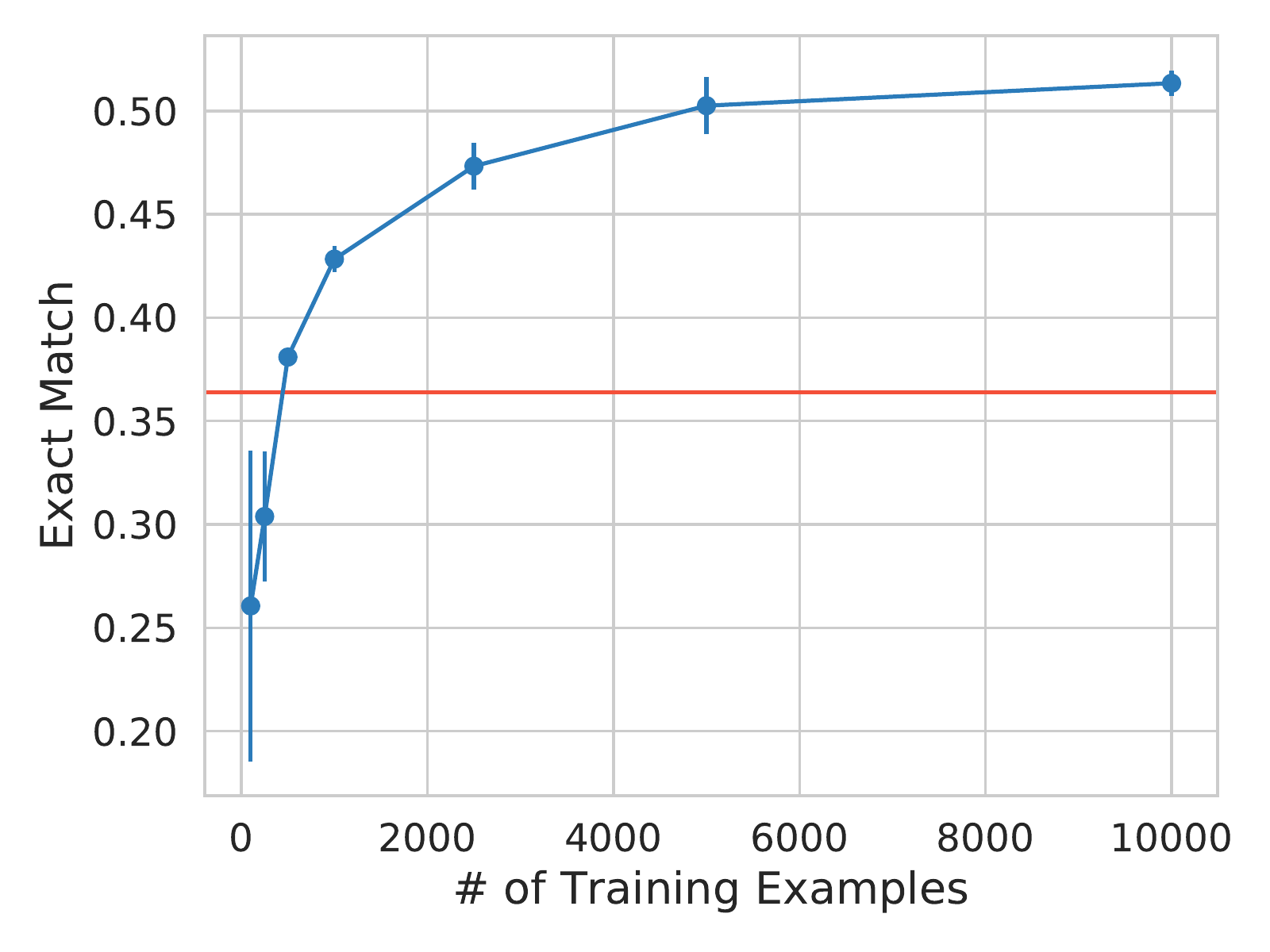}
\caption{Results from T5\label{fig:size_t5}}
\end{subfigure}
\caption{Size Experiments on HotpotQA. The red line indicates the performance of zero-shot decompositions while the blue line is the No Decomposition method.\label{fig:size}
\vspace{-1em}}
\end{figure}

\paragraph{Size Experiment} How much data is needed for the \textit{No Decomposition} method with fine-tuning to perform comparably to decomposition methods without fine-tuning (e.g. the setting in \citet{patel2022question})? We answer this question in Figure~\ref{fig:size}. As we need a model fine-tuned on QA to evaluate in the zero-shot case, we use UnifiedQA-v2 and a version of T5 fine-tuned on SQuAD \cite{rajpurkar2016squad}.\footnote{Note that we do not evaluate on DROP as UnifiedQA-v2 was already trained on DROP.} Points indicate the mean score over three random seeds and error bars indicate one standard deviation. We see that as the amount of data increases, the fine-tuned model performs better; it is better than the zero-shot method between 100-250 examples for UnifiedQA-v2 and 250-500 for T5.

\paragraph{Error Analysis}
Why does the standard not fine-tuned decomposition method (e.g. \citet{patel2022question}) perform worse than the fine-tuned \textit{No Decomp} method?
In Table~\ref{table:error analysis} we show representative errors from the decomposition approach with how often they occurred. We randomly sample 50 instances of errors and categorize them into three groups: wrong predictions in the last step, error propagation from intermediate steps, and invalid/missing annotations from BREAK (i.e. not the model's fault). We found that the biggest category was predicting an invalid annotation (42\%), i.e. an alias that the dataset did not contain, followed by error propagation (40\%) and then wrong predictions (18\%). Thus, compared to the non-iterative methods, the iterative process allows error propagation that occurs in roughly 40\% of errors, contributing to its lower comparative performance.

\section{Related Work}
\paragraph{Decompositions in QA}
Decompositions for QA have a long history in complex question answering \cite {perez2020unsupervised,geva2021did} with recent interest in using them for large language models \cite{wei2022chain,dua2022successive}. Two of the most related works to ours include \citet{patel2022question} who show improvements when using decompositions in the zero-shot setting and concurrent work \cite{guo2022complex} that uses decompositions on the full DROP dataset (beyond BREAK). Note that our reported no-decomposition results on DROP show comparable or greater performance to \citet{guo2022complex} when using only the BREAK dataset. Our work complements these papers by showing when decompositions help w.r.t. data size.

\paragraph{Decompositions in other fields}
Our understanding of decompositions in textual question answering helps to unify results across machine learning fields, as similar results have been shown in Computer Vision \cite{hudson2019gqa,amizadeh2020neuro,li2021calibrating} and through results on similar visual QA leaderboards.

\paragraph{Decomposition Strategies and Prompting}
Decompositions methods are also related to prompting, where the explicit decompositions can be seen as a hard prompt \cite{liu2021pre,su2022transferability} and the implicit decompositions are similar to soft prompts \cite{qin-eisner-2021-learning,liu2022p}.

\section{Conclusion}
Our work explored when decompositions are helpful for machine reading. We showed that decompositions are helpful when there is limited data available, or when parameters cannot be tuned. However, when enough data exists (empirically around a few hundred instances) and parameters can be fine-tuned, it is best to let the model learn the decompositions implicitly through end-to-end training. 
We hope that our work will help to inform readers as they create new datasets and select methods to use for complex question answering.

\bibliography{anthology,custom}
\bibliographystyle{acl_natbib}

\appendix

\section{Hyperparameter and Compute Details}
\label{app:details}
Models were trained with a default learning rate of 1e-5 for 5 epochs, using early stopping on a holdout of the training set (5\%) to determine the best saved model. The best models were typically ones trained for around 2 epochs. 

For the non-fine-tuned decomposition UnifiedQA-v2 approach (perhaps due to its multi-task pre-training on other QA datasets), we found that using the question plus the iterative decompositions added a solid boost to final performance and hence we use those results. Each non-fine-tuned decomposition run takes approximately 15-30 minutes to evaluate.

Compute time ranged from 1 hr for the shortest jobs with smaller data sizes and non-iterative training to 18 hours for iterative decompositions methods with UnifiedQA-v2 on 1 RTX 6000 GPU. 

Data was prepared using the original BREAK authors' code. Models were accessed via the Huggingface repository \cite{wolf2019huggingface}.

\end{document}